# Rhyme-aware Chinese lyric generator based on GPT

Yixiao Yuan*[a], Yangchen Huang[a], Yu Ma[b], Xinjin Li[a], Zhenglin Li[c], Yiming Shi[d], Huapeng Zhou[e]
[a]Columbia University, New York, NY, USA
yixiao.yuan@columbia.edu, huangyangchen88@gmail.com, li.xinjin@columbia.edu
[b]Carnegie Mellon University, Pittsburgh, PA, USA
yuma13926@gmail.com
[c]Texas A&M University, College Station, TX, USA
zhenglin_li@tamu.edu
[d]Xi'an Jiaotong-Liverpool University, Suzhou, China
yiming.shi21@student.xjtlu.edu.cn
[e]University of Washington, Seattle, WA, USA
zhouhp.me@gmail.com

## ABSTRACT

Neural language representation models such as GPT, pre-trained on large-scale corpora, can effectively capture rich semantic patterns from plain text and be fine-tuned to consistently improve natural language generation performance. However, existing pre-trained language models used to generate lyrics rarely consider rhyme information, which is crucial in lyrics. Using a pre-trained model directly results in poor performance. To enhance the rhyming quality of generated lyrics, we incorporate integrated rhyme information into our model, thereby improving lyric generation performance.

## 1. INTRODUCTION

Writing lyrics is challenging, even for experienced lyricists. We aim to design an AI lyrics generator to inspire lyricists when they lack inspiration. To the best of our knowledge, existing methods for Chinese lyrics are inadequate, prompting us to seek modifications for improved results. We believe this work is both interesting and valuable.

Pre-trained models[1,2,3,4] have achieved enormous success in various natural language processing tasks such as machine translation, information retrieval, and text generation. These models can learn extensively from large corpora, allowing them to represent sentence semantics effectively. In this project, we utilize a pre-trained model to aid in the generation of Chinese lyrics.

Traditional natural language generation methods do not require rhyme, resulting in poor-quality lyrics when these methods are used. Ernie[5] integrates entity representations in its knowledge module, learning a substantial amount of general knowledge from knowledge graphs. Inspired by Ernie, we decided to incorporate integrated rhyme information into our model to help it learn the rhyme of Chinese characters.

Our work has two main contributions. First, we successfully added rhyme embedding to the GPT-2 model to generate better rhyming lyrics. Second, we propose a new way to integrate the rhyme of Chinese characters and word representation. In our self-made test dataset, we achieved an 82.2% rhyme ratio, while the pre-trained model only achieved a 30.9% rhyme ratio.

## 2. METHOD

### 2.1 Model architecture

As shown in Figure 1, the whole model architecture of our work consists of two stacked modules. The underlying textual encoder responsible to capture basic lexical and syntactic information from the input tokens, and the upper encoder responsible to integrate extra rhyme information into textual information from the underlying layer. Therefore, we can represent heterogeneous information of tokens and entities into a united feature space.

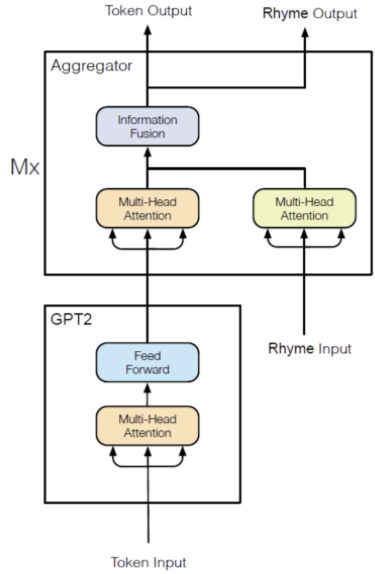

Figure 1. Model architecture. Token is encoded by pre-trained GPT, and token and rhyme input are fusion by aggregator.

## 2.2 Rhyme vocabulary

To obtain rhyme embeddings, the first step is to create the rhyme vocabulary. We manually classified pinyin into 13 classes; for example, "ai" and "uai" are placed in the same class. Rhymes within the same class have similar pronunciations. Additionally, we include four tags: <pad>, <cls>, <sep>, and <space>, where <space> represents a pause in a sentence. For English words or other characters that cannot be represented with pinyin, we mark them as <unknown> for convenience. This results in a one-hot label of length 18. After the one-hot input, we use a linear layer to obtain a dense representation. The rhyme vocabulary are shown in the Table 1.

Table 1. Rhyme with the same class. The first line stand for four speicial tags. All the rhyme in the same class is listed in the same line.

| <pad> | <cls> | <seq> | <space> | | |
|---|---|---|---|---|---|
| a | ia | ua | | | |
| an | ian | uan | | | |
| o | e | uo | | | |
| en | un | | | | |
| ie | ve | | | | |
| ang | iang | uang | | | |
| ai | uai | | | | |
| eng | ong | iong | | | |
| ei | ui | | | | |
| i | er | v | in | ing | |
| ao | Iao | | | | |
| ou | Iu | | | | |
| u | | | | | |

## 2.3 Rhyme-aware encoder

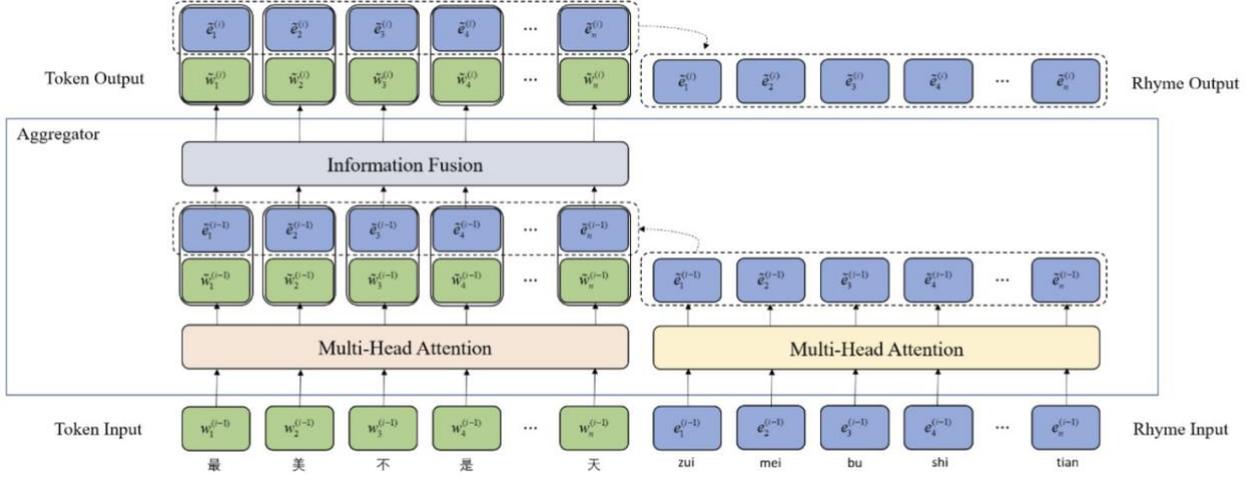

Figure 2. Aggregator architecture. Information fusion layer takes two kinds of input: one is the token embedding, and the other one is the concatenation of the token embedding and rhyme embedding. After information fusion, it outputs new token embeddings and rhyme embeddings for the next layer. Token Input: Chinese characters (e.g., '最美不是…天' meaning 'The most beautiful is not the sky') Rhyme Input: Corresponding pinyin endings (e.g., 'zui mei bu shi … tian')

As shown in Figure 2, the rhyme-aware encoder consists of stacked aggregators, which are designed for encoding both tokens and rhyme as well as fusing their heterogeneous features. In the i-th aggregator, the input token embeddings $\{w_1^{(i-1)}, \ldots, w_n^{(i-1)}\}$ and rhyme embeddings $\{e_1^{(i-1)}, \ldots, e_n^{(i-1)}\}$ from the preceding aggregator are fed into two mask multi-head self-attentions (MMH-ATTs)[6], respectively. This is to make the model realize the last word in the previous sentence.

$$\{\widetilde{w_1^{(i)}}, \ldots, \widetilde{w_n^{(i)}}\} = \text{MMH-ATT}\left(\{w_1^{(i-1)}, \ldots, w_n^{(i-1)}\}\right)$$

$$\{e_1^{(i)}, \ldots, e_m^{(i)}\} = \text{MMH-ATT}\left(\{e_1^{(i-1)}, \ldots, e_m^{(i-1)}\}\right)$$

Then, the i-th aggregator adopts an information fusion layer for the mutual integration of the token and rhyme, and computes the output embedding for each token and rhyme. For a token $w_j$ and its rhyme $e_j$ the information fusion process is as follows,

$$h_j = \sigma\left(\widetilde{W_t^{(i)}}\widetilde{w_j^{(i)}} + \widetilde{W_e^{(i)}}\widetilde{e_k^{(i)}} + \widetilde{b^{(i)}}\right)$$
$$w_j^{(i)} = \sigma(W_t^{(i)} h_j + b_t^{(i)})$$
$$e_k^{(i)} = \sigma(W_e^{(i)} h_j + b_e^{(i)})$$

Additionally, we incorporate layer normalization and residual connections between the two aggregators. We compared models with and without these layers, finding that residual connections help the model converge. After testing for 40 epochs, we observed that the loss for the model without these layers fluctuates around 6, whereas the model with these layers sees a continuous decline in loss, reaching 2.5 with a continued downward trend. Clearly, residual connections enhance our model's performance.

## 3. EXPERIMENTS

### 3.1 Dataset and pre-processing

The Chinese-Lyric-Corpus is designed for the task of Chinese lyric generation, containing nearly 50,000 Chinese lyrics from 500 artists. To better learn the rhyme information between sentences, we train the model using every two sentences in a song instead of a single lyric. This approach helps the model capture the rhyme of the last few words.

We create two datasets to train the model. To enhance the model's ability to extract useful information, we filter some sentences. For instance, many lyrics contain repeated lines with the same rhyme. However, we want the model to generate richer content. Therefore, we remove sentence pairs where the end word is the same. Additionally, we set a threshold for the length of lyrics to avoid collecting onomatopoeic expressions, which may be classified as sentences during data processing, such as "啊" and "呀". This results in our first dataset, which we call the Filtered dataset.

Furthermore, we use only rhymed sentences as our training data to obtain more rhyming results, which obviously increases the model's performance. This is our second dataset, referred to as Only Rhyme. The results and further details will be discussed in Section 3.3.

### 3.2 Implementation details

The proposed network is implemented using PyTorch and is trained on a single NVIDIA RTX 3090 GPU. The model undergoes training for 40 epochs using the Adam optimizer, with an initial learning rate of 1e-5 and a warm-up period of 4 epochs. After reaching the default learning rate, it decays linearly. The loss function used is cross-entropy loss. The model comprises 6 aggregators with a fusion dimension of 768, and each aggregator employs a masked attention mechanism with 4 heads. The batch size for training is set to 64, and gradient accumulation is used for 2 steps.

### 3.3 Ablation studies

To demonstrate the effectiveness of our proposed method, we compared it with a model that does not use rhyme input. All parameters were kept the same, with the only difference being the removal of rhyme input. Given the complexity of lyric writing, we conducted a human evaluation to assess the performance of different models. Each lyric was assessed by three individuals in a blind review manner, where the reviewers had no information about the generation method used for each lyric. Following previous work in generating poems[7,8], we evaluated the generated lyrics based on three criteria: consistency, fluency, and meaning. Each criterion was rated from 1 to 3, representing bad, normal, and good, respectively. Additionally, we added a rhyming rate to evaluate rhyming performance. The details are illustrated in Table 3.

As shown in the table, our rhyme embedding method is very effective compared to methods without rhyme embedding. Using raw data, the rhyme embedding increases the rhyming rate by 10 percent. With processed data, it increases the rhyming rate by 20 percent. Furthermore, comparing raw and processed data, the model using processed data doubles the rhyming rate.

This result is easily explained. Rhyme embedding incorporates rhyme information into the model, enabling it to learn more about rhyming. About half of the original data is rhyming, so the model can still learn to rhyme without processing data. After processing, all the training data is rhyming, allowing our method to achieve a rhyming rate of 0.822. It is noteworthy that we did not use any rhyme-specific loss during training, indicating that our model learned rhyming information in lyrics without any supervision. However, we did not compare adding rhyme-specific loss to adding rhyme embedding. We suggest that adding rhyme-specific loss might cause the model to lose the ability to learn semantic information in lyrics and generate rhyming but meaningless sentences. Additionally, the rhythmic beauty of lyrics is reflected not only at the end of sentences but throughout the entire sentence. Rhyme-specific loss can only learn rhyme at the end of sentences, whereas rhyme embedding can capture more comprehensive rhyme information, even if it cannot be clearly explained.

In our method, the model learns rhyme information without any supervision, which we believe allows it to learn rhyming information without diminishing the meaning. The results support our suggestion. The metrics for consistency, fluency, and meaning are comparable to those of methods without rhyme embedding. However, since we relied on human evaluation, the variance is large.

We also present some results generated by the two models, showing that our model with rhyme input outperforms the other in both qualitative and quantitative analyses.

Table 3. Information on video and audio files that can accompany a manuscript submission.

| Methods | Rhyming rate | Consistency | Fluency | Meaning |
| --- | --- | --- | --- | --- |
| w/o rhyming embedding | 0.309 | 2 | 1.33 | 2 |
| w/ rhyming embedding | 0.407 | 2.33 | 1.67 | 1.67 |
| w/o rhyming embedding + processed data | 0.618 | 2 | 1.67 | 2.33 |
| w/ rhyming embedding + processed data | **0.822** | 2 | 2.33 | 2 |

## 3.4 Result

Here are the results generated by our method. We found that many of the generated lyrics focus on love, which reflects the love-centric nature of our training data. To reduce the prevalence of love-themed lyrics, we need a more diverse dataset.

In Figure 3, the generated sentences on the left all use the rhyme "an," showing that our method produces rhymed lyrics. The generated lyrics appear meaningful and can be hard to distinguish from human-made ones at first glance.

Not all sentences rhyme perfectly; on the right, two sentences have different rhymes. However, the overall lyric maintains coherence, with recurring themes like "雨" demonstrating consistency. This coherence is due to the self-attention mechanism in the transformer, which handles long-distance dependencies effectively.

In summary, our method effectively generates rhymed and meaningful lyrics, with the self-attention module enhancing content coherence. Expanding the training dataset can further improve the variety and quality of the lyrics.

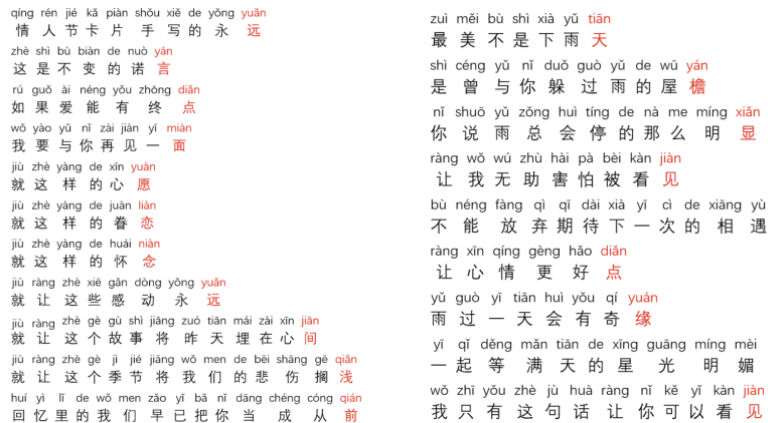

Figure 3. Two lyrics generated by our model. All the rhymed words at the end of every sentence are labeled in red. In these lyrics, there are two sentences that have end words with different rhymes. The left lyric is about eternal love and memories. The right lyric is about rain and hope.

## 3.5 Attention visualization

To check if our model works well, we visualized the attention in Transformer models. In the visualization, darker colors indicate greater contribution.

Attention visualization in the token layer (left side of Figure 4) shows that every word contributes significantly to the next, with attention to both nearby and distant words, ensuring consistency. In the rhyme layer visualization (right side of Figure 4), rhymes within the same class show a strong contribution to each other. For example, the end of "天" greatly influences the end of "现", explaining why our model generates more rhyming lyrics.

These visualizations demonstrate that our model effectively captures both local and global dependencies in lyrics, contributing to the generation of coherent and rhyming lyrics.

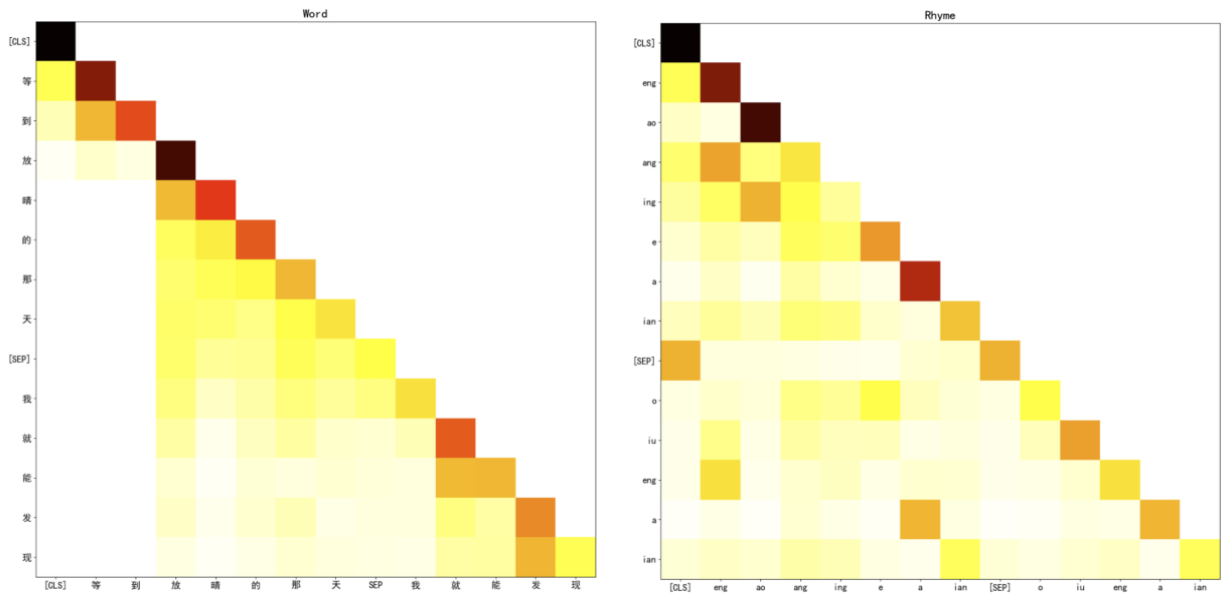

Figure 4. Attention visualization in word and rhyme layers. The darker the color, the greater the contribution. Left: Word layer attention. The vertical axis shows Chinese characters, with their corresponding pinyin (romanization) on the horizontal axis. Right: Rhyme layer attention. Both axes display pinyin endings, representing rhyme sounds. [CLS] and [SEP] are special tokens used in the model architecture.

## 4. CONCLUSION

In this paper, we propose a model called the Rhyme-Aware Chinese Lyric Generator, which incorporates rhyme information and is based on GPT-2. Experimental results show that our method outperforms models without rhyme embedding in generating lyrics. However, our work has limitations. The training data is not large enough, and we only consider rhyme without addressing intonation, which is also important in lyrics. Additionally, expanding the dataset and incorporating more prior knowledge could further enhance the model's performance in limited data scenarios.